\documentclass[11pt,a4paper]{article}
\usepackage[hyperref]{acl2021}
\usepackage{times}
\usepackage{latexsym}

\usepackage{microtype}
\usepackage{relsize}
\usepackage{booktabs}
\usepackage{multirow}
\usepackage{listings}
\usepackage{tikz}
\usepackage{pgfplots}
\usepackage{amsmath}
\usepackage{tikz-dependency}
\usetikzlibrary{patterns,calc,shapes,backgrounds,shadows,positioning,fit,matrix,shapes.geometric}
\usetikzlibrary{automata, arrows}
\usepackage{xsavebox}
\usepackage{array}
\newcommand{\PreserveBackslash}[1]{\let\temp=\\#1\let\\=\temp}
\newcolumntype{C}[1]{>{\PreserveBackslash\centering}p{#1}}
\newcolumntype{R}[1]{>{\PreserveBackslash\raggedleft}p{#1}}
\newcolumntype{L}[1]{>{\PreserveBackslash\raggedright}p{#1}}

\aclfinalcopy 
\def\aclpaperid{9} 

\title{One Semantic Parser to Parse Them All: Sequence to Sequence Multi-Task Learning on Semantic Parsing Datasets}

\author{Marco Damonte \quad Emilio Monti \\
  Amazon Alexa AI \\
  {\tt \{dammarco,monti\}@amazon.com}}
\date{}

\begin{document}
\maketitle

\begin{abstract}
Semantic parsers map natural language utterances to meaning representations. The lack of a single standard for meaning representations led to the creation of a plethora of semantic parsing datasets. To unify different datasets and train a single model for them, we investigate the use of Multi-Task Learning (MTL) architectures. We experiment with five datasets (\textsc{Geoquery}, \textsc{NLMaps}, \textsc{TOP}, \textsc{Overnight}, \textsc{AMR}).
We find that an MTL architecture that shares the entire network across datasets yields competitive or better parsing accuracies than the single-task baselines, while reducing the total number of parameters by 68\%. We further provide evidence that MTL has also better compositional generalization than single-task models. We also present a comparison of task sampling methods and propose a competitive alternative to widespread proportional sampling strategies.

\end{abstract}

\section{Introduction}
\label{intro}

Semantic parsing is the task of converting natural language into a meaning representation language (MRL). The commercial success of personal assistants, that are required to understand language, has contributed to a growing interest in semantic parsing. A typical use case for personal assistants is Question Answering (Q\&A): the output of a semantic parser is a data structure that represents the underlying meaning of a given question. This data structure can be compiled into a query to retrieve the correct answer. 
The lack of a single standard for meaning representations resulted in the creation of a plethora of semantic parsing datasets, which differ in size, domain, style, complexity, and in the formalism used as an MRL. These datasets are expensive to create, as they normally require expert annotators. Consequently, the datasets are often limited in size.

Multi-task Learning (MTL; \citealt{caruana1997multitask}) refers to jointly learning several tasks while sharing parameters between them. In this paper, we use MTL to demonstrate that it is possible to unify these smaller datasets together to train a single model that can be used to parse sentences in any of the MRLs that appear in the data. We experiment with several Q\&A semantic parsing dataset for English: \textsc{Geoquery} \citep{zelle1996learning}, \textsc{NLMaps v2} \citep{lawrence2018improving}, \textsc{TOP} \citep{gupta2018semantic}, and \textsc{Overnight} \citep{wang2015building}. 
In order to investigate the impact of less related tasks, we also experiment on a non-Q\&A  semantic parsing dataset, targeting a broader coverage meaning representation: \textsc{AMR} \citep{Banarescu13abstractmeaning}, which contains sentences from sources such as broadcasts, newswire, and discussion forums.

Our baseline parsing architecture is a reimplementation of the sequence to sequence model by \newcite{rongali2020don}, which can be applied to any parsing task as long as the MRL can be expressed as a sequence. Inspired by \newcite{fan2017transfer}, we experimented with two MTL architectures: \textsc{1-to-N}, where we share the encoder but not the decoder, and \textsc{1-to-1}, where we share the entire network. Previous work \citep{ruder2017overview,collobert2008unified,hershcovich2018multitask} has focussed on a lesser degree of sharing more closely resembling the \textsc{1-to-N} architecture, but we found \textsc{1-to-1} to consistently work better in our experiments. 

In this paper we demonstrate that the \textsc{1-to-1} architecture can be used to achieve competitive parsing accuracies for our heterogeneous set of semantic parsing datasets, while reducing the total number of parameters by 68\%, overfitting less, and improving on a compositional generalization benchmark \cite{keysers2019measuring}.

We further perform an extensive analysis of alternative strategies to sample tasks during training. A number of methods to sample tasks proportionally to data sizes have been recently proposed \cite{wang2019glue,sanh2019hierarchical,wang2019can,stickland2019bert}, which are often used as de facto standards for sampling strategies. These methods rely on the hypothesis that sampling proportionally to the task sizes avoids overfitting the smaller tasks. We show that this hypothesis is not generally verified by comparing proportional methods with an inversely proportional sampling method, and a method based on the per-task loss during training. Our comparison shows that there is not a method that is consistently superior to the others across architectures and datasets. We argue that the sampling method should be chosen as another hyper-parameter of the model, specific to a problem and a training setup.

We finally run experiments on dataset pairs, resulting in 40 distinct settings, to investigate which datasets are most helpful to others. Surprisingly, we observe that \textsc{AMR} and \textsc{Geoquery} can work well as auxiliary tasks. 
\textsc{AMR} is the only graph-structured, non Q\&A dataset, and was therefore not expected to help as much as more related Q\&A datasets. \textsc{Geoquery} is the smallest dataset we tested, showing that low-resource datasets can help high-resource ones instead of, more intuitively, the other way around.


\section{Sequence to Sequence Multi-Task Learning}
\label{sec:seq2seq_mtl}

\begingroup\def\pdfdest name#1#2{} 
\begin{xlrbox}{NLMAPS}
\begin{minipage}{0.35\textwidth}
{\small
  \begin{lstlisting}[basicstyle=\tiny]
query(
  area(
    keyval('name','Edinburgh')), 
  nwr(
    keyval('tourism', 'hotel')), 
  qtype(count))
  ...
\end{lstlisting}
}
\end{minipage}
\end{xlrbox}
\begin{xlrbox}{GEOQUERY}
\begin{minipage}{0.35\textwidth}
  \begin{lstlisting}[basicstyle=\tiny]
answer(
  count(
    hotel(
      loc_2(
        city_id('Edinburgh'))))
        ...
\end{lstlisting} 
\end{minipage} 
\end{xlrbox}
\begin{figure*}
 \centering
  \begin{tikzpicture}
    \draw (0,0) node(utt) {\small \emph{Number of hotels in Edinburgh}};
    \draw (3.5,0) node(e) [rectangle,draw,very thick, minimum width=2cm, minimum height=0.8cm] {Encoder};
    \draw (6,1) node(d1) [rectangle,draw,very thick, minimum width=3cm, minimum height=0.8cm] {\small{Decoder \textsc{NLMaps}}};
    \draw (6,-1) node(d2) [rectangle,draw,very thick, minimum width=3cm, minimum height=0.8cm] {\small{Decoder \textsc{Geoquery}}};
    \draw (10.7,1) node(mrl1) {\color{blue}\theNLMAPS};
    \draw (10.7,-1) node(mrl2) {\color{blue}\theGEOQUERY};
    \draw [->,very thick] (utt) -- node[]{} (e);
    \draw [->,very thick] (e) -- node[]{} (d1);
    \draw [->,very thick] (e) -- node[]{} (d2);
    \draw [->,very thick] (d1) -- node[above]{} (mrl1);
    \draw [->,very thick] (d2) -- node[above]{} (mrl2);

    \draw (0,-2) node(utt1) {\small \emph{$<$\textsc{NLMaps}$>$ Number of hotels in Edinburgh}};
    \draw (0,-5.4) node(utt2) {\small \emph{$<$\textsc{Geoquery}$>$ Number of hotels in Edinburgh}};
    \draw (3.5,-4.2) node(e) [rectangle,draw,very thick, minimum width=2cm, minimum height=0.8cm] {Encoder};
    \draw (6,-4.2) node(d1) [rectangle,draw,very thick, minimum width=2cm, minimum height=0.8cm] {Decoder};
    \draw (10.5,-3.5) node(mrl1) {\color{blue}\theNLMAPS};
    \draw (10.5,-5.1) node(mrl2) {\color{blue}\theGEOQUERY};
    \draw [->,very thick] (utt1) -- node[]{} (e);
    \draw [->,very thick] (utt2) -- node[]{} (e);
    \draw [->,very thick] (e) -- node[]{} (d1);
    \draw [->,very thick] (d1.east) -- node[above=0.3cm]{} (mrl1.west);
    \draw [->,very thick] (d1.east) -- node[below=0.3cm]{} (mrl2.west);
  \end{tikzpicture}  
 \caption{Two MTL architectures for two tasks (A and B): at the top \textsc{1-to-N}, where only the encoder is shared; at the bottom \textsc{1-to-1}, where we also share the decoder and we add a special token at the beginning of the input sentence.}
 \label{fig:mtl_setups}
\end{figure*}
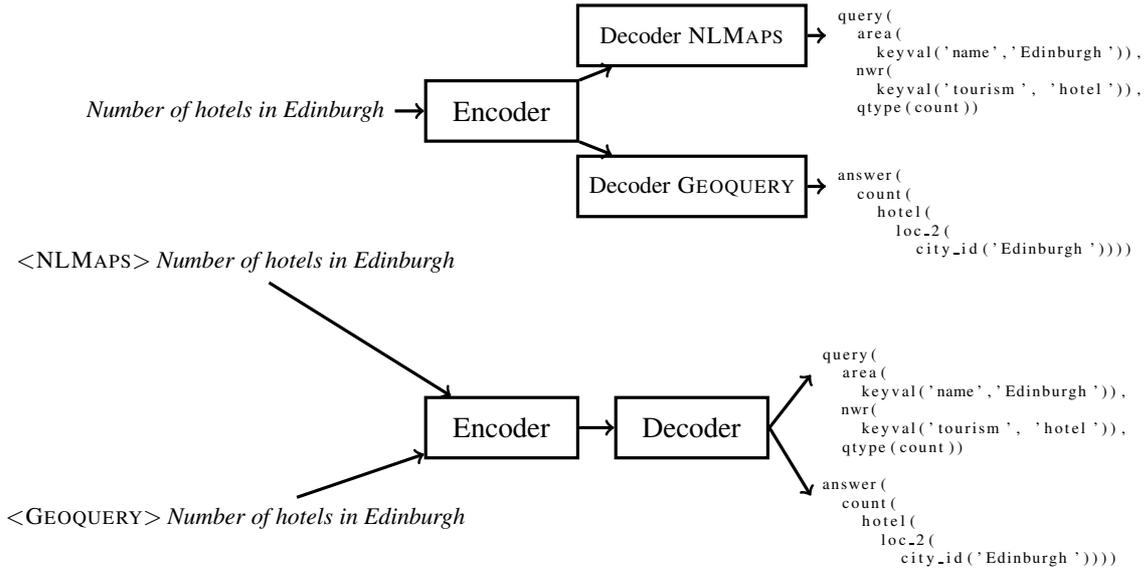
\endgroup

MTL refers to machine learning models that sample training examples from multiple tasks and share parameters amongst them. During training, a batch is sampled from one of the tasks and the parameter update only impacts the part of the network relevant to that task. 

The architecture for sequence to sequence semantic parsing that we use in this paper consists of an encoder, which converts the input sentence into a latent representation, and a decoder, which converts the latent representation into the output MRL \citep{jia2016data,konstas2017neural,rongali2020don}. While the input to each task is always natural language utterances, each task is in general characterized by a different meaning representation formalism. It, therefore, follows that the input (natural language) varies considerably less than the output (the meaning representation). Parameter sharing can therefore more intuitively happen in the encoder, where we learn parameters that encode a representation of the natural language. Nevertheless, more sharing can also be allowed, by also sharing parts of the decoder \citep{fan2017transfer}. In this work, we experiment with two MTL architectures, as shown in Figure~\ref{fig:mtl_setups}: \textsc{1-to-N}, where we share the encoder but not the decoder, and \textsc{1-to-1}, where we share the entire network. As different datasets normally use different MRLs, in the \textsc{1-to-1} architecture we also need a mechanism to inform the network of which MRL to generate. We therefore augment the input with a special token that identifies the task, following \newcite{johnson2017google}.



\section{Experimental Setup}
In this section, we describe the datasets used, baseline architectures, and training details.

\subsection{Data}
\label{sec:data}
\begingroup\def\pdfdest name#1#2{}
\begin{xlrbox}{EXAMPLEGEOQUERY}
\begin{minipage}{0.6\textwidth}
  \begin{lstlisting}
answer(
  shortest(
    river(all)))
    ... 
\end{lstlisting} 
\end{minipage} 
\end{xlrbox}
\begin{xlrbox}{EXAMPLETOP}
\begin{minipage}{0.6\textwidth}
  \begin{lstlisting}
[IN:GET_INFO_TRAFFIC 
  is traffic heavy in 
  [SL:LOCATION
    downtown]]
    ...
\end{lstlisting} 
\end{minipage} 
\end{xlrbox}
\begin{xlrbox}{EXAMPLENLMAPS}
\begin{minipage}{0.6\textwidth}
  \begin{lstlisting}
query(
  area(keyval('name','Nantes')), 
  nwr(keyval('place','locality')), 
  qtype(findkey('name')))
  ...
\end{lstlisting} 
\end{minipage} 
\end{xlrbox}
\begin{xlrbox}{EXAMPLEOVERNIGHT}
\begin{minipage}{0.6\textwidth}
  \begin{lstlisting}
listValue(
  filter(
    filter(getProperty(
        singleton en.meeting) 
      (string !type)) 
    (string is\_important)))
    ...

\end{lstlisting}
\end{minipage} 
\end{xlrbox}
\begin{xlrbox}{EXAMPLEAMR}
\begin{minipage}{0.6\textwidth}
  \begin{lstlisting}
(p / pollute-01 :polarity -
  :ARG0 (m / method 
    :mod (t / this)) 
  :ARG1 (e / environment))
  ...
\end{lstlisting} 
\end{minipage} 
\end{xlrbox}
\begin{table*}
\centering
\begin{tabular}{lll}
\toprule
\textbf{Dataset} & \textbf{Input} & \textbf{Output}\\
\textsc{Geoquery} & which is the shortest river & \theEXAMPLEGEOQUERY\\
\midrule
\textsc{NLMaps} & name Localities in Nantes & \theEXAMPLENLMAPS\\
\midrule
\textsc{TOP} & is traffic heavy downtown & \theEXAMPLETOP\\
\midrule
\textsc{Overnight} & show me all & \multirow{2}{*}{\theEXAMPLEOVERNIGHT}\\ & important meetings & \\
& & \\
& & \\
& & \\
& & \\
\midrule
\textsc{AMR} & this method will not & \multirow{2}{*}{\theEXAMPLEAMR}\\ & pollute the environment & \\
& & \\
& & \\
& & \\
\bottomrule
\end{tabular}
\caption{Training examples from each of the datasets used in our experiments. The output logical forms were simplified for the sake of readability.}
\label{tab:data_examples}
\end{table*}
\endgroup
\begin{table*}
\centering
\begin{tabular}{lccccc}
\toprule
\textbf{Dataset} & \textbf{Train} & \textbf{Dev} & \textbf{Test} & \textbf{Src Vocab} & \textbf{Tgt Vocab}\\
\midrule
\textsc{Geoquery} & 540 & 60 & 280 &  279 & 103 \\
\textsc{NLMaps} & 16172 & 1843 & 10594 & 8628 & 1012 \\
\textsc{TOP} & 28414 & 4032 & 8241 & 11873 & 116 \\
\textsc{Overnight} & 18781 & 2093 & 5224 & 1921 & 311 \\
\textsc{AMR} & 36521 & 1368 & 1371 & 30169 & 28880 \\
\bottomrule
\end{tabular}
\caption{Details of each dataset. ``Train'', ``Dev'', and ``Test'' are the number of examples (questions paired with MRLs) in the training, development, and test splits. ``Src Vocab'' is the vocabulary size for the input (natural language) and ``Tgt Vocab'' is the vocabulary size for the output (meaning representation).}
\label{tab:stats}
\end{table*}

While we focussed on Q\&A semantic parsing datasets, we further consider the \textsc{AMR} dataset in order to investigate the impact of MTL between considerably different datasets. Table~\ref{tab:data_examples} shows a training example from each dataset. The sizes of all datasets are shown in Table~\ref{tab:stats}. 

\paragraph{Geoquery} Questions and queries about US geography \citep{zelle1996learning}. The best results on this dataset are reported by \newcite{kwiatkowski2013scaling} via Combinatory Categorial Grammar \citep{surfacesteedman,syntaxsteedman} parsing. 

\paragraph{NLMaps v2} Questions about geographical facts \citep{lawrence2018improving}, retrieved from OpenStreetMap \citep{haklay2008openstreetmap}. To our knowledge, we are the first to train a parser on the full dataset. Previous work trained a neural parser on a small subset of the dataset and used the rest to experiment with feedback data \citep{lawrence2018counterfactual}. 
We note that there exists a previous version of the dataset \citep{haas2016corpus}, for which state-of-the-art results have been achieved with a sequence to sequence approach \citep{duong2017multilingual}. We use the latest version of the dataset due to its larger size. 

\paragraph{TOP} Navigation and event queries generated by crowdsourced workers \citep{gupta2018semantic}. The queries are annotated to semantic frames comprising of intents and slots. The best results are achieved by a sequence to sequence model \citep{aghajanyan2020conversational}. 

\paragraph{Overnight} This dataset \citep{wang2015building} contains Lambda DCS \citep{liang2013lambda} annotations divided into eight domains: \emph{calendar}, \emph{blocks}, \emph{housing}, \emph{restaurants}, \emph{publications}, \emph{recipes}, \emph{socialnetwork}, and \emph{basketball}. Due to the small size of the domains, we merged them together. The current state-of-the-art results, on single domains, are reported by \newcite{su2017crossdomain}, who frame the problem as a paraphrasing task. They use denotation (answer) accuracy as a metric, while we report parsing accuracies, a stricter metric.

\paragraph{AMR} AMR \cite{Banarescu13abstractmeaning} has been widely adopted in the semantic parsing community \citep{artzi2009broad,carbonell2014discriminative,wang,damonte2017incremental,titov2007latent,zhang2019broad}. We used the latest version of the dataset (LDC2017T10), for which the best results were reported by \newcite{bevilacquaone}.
The \textsc{AMR} dataset is different from the other datasets, not only in that it is not Q\&A, but also in the formalism used to express the meaning representations. While for the other datasets the output logical forms can be represented as trees, in \textsc{AMR} each sentence is annotated as a rooted, directed graph, due to explicit representation of pronominal coreference, coordination, and control structures. 

In order to use sequence to sequence architectures on \textsc{AMR}, a preprocessing step is required to remove variables in the annotations and linearize the graphs. In this work, we followed the linearization method by \newcite{van2017dealing}.\footnote{\url{https://github.com/RikVN/AMR with default settings}}

\subsection{Baseline Parser}

Our baseline parser is a reimplementation of \newcite{rongali2020don}: a single-task attentive sequence to sequence model \cite{bahdanau2015neural} with pointer network \citep{vinyals2015pointer}. The input utterance is embedded with a pretrained \textsc{RoBERTa} encoder \cite{liu2019roberta}, and subsequently fed into a \textsc{Transformer} \cite{vaswani2017attention} decoder. 
The encoder converts the input sequence of tokens $x_1, \dotsc, x_n$  into a sequence of context-sensitive embeddings $e_1, \dotsc, e_n$. 
At each time step $t$, the decoder generates an action $a_t$. There are two types of actions: output a symbol from the output vocabulary, or output a pointer to one of the input tokens $x_i$. The final softmax layer provides a probability distribution, for $a_t$, across all these possible actions. The probability with which we output a pointer to $x_i$ is determined by the attention score on $x_i$. Finally, we use beam search to find the sequence of actions that maximize the overall output sequence probability.

\subsection{Training}
\label{sec:hyper}

All models were trained with Adam \citep{kingma2014adam} on P3 AWS machines with one Tesla V100 GPU. To prevent overfitting, we used an early stopping policy to terminate training once the loss on the development set stops decreasing. To account for the effect of the random seed used for initialization, we train three instances of each model with different random seeds. We then report the average and standard deviation on the test set. 

We evaluate all Q\&A parsing models using the exact match metric, which is computed as the percentage of input sentences that are parsed without any mistake. \textsc{AMR} is instead evaluated using \textsc{Smatch} \citep{cai2013smatch}, which computes the F1 score of graphs' nodes and edges.\footnote{\url{https://github.com/snowblink14/smatch}} 

We tuned hyper-parameters for each model based on exact match accuracies on their development sets. While \textsc{AMR} is typically evaluated on \textsc{Smatch}, to simplify the tuning of our models, we use exact match also for \textsc{AMR} and compute the \textsc{Smatch} score only for the final models. 
We performed manual searches (5 trials) for the following hyper-parameters: batch size (10 to 200), learning rate (0.04 to 0.08), number of layers (2 to 6) and units in the decoder (256 to 1024), number of attention heads (1 to 16), and dropout ratio (0.03 to 0.3). 
For the baseline, we selected the sets of hyper-parameters that maximize performance on the development set of each dataset. To tune the MTL model for each dataset would be costly: we instead selected the set of parameters that maximizes performance on the combination of all development sets. For analogous reasons, when presenting results on MTL between the 40 combinations of dataset pairs, we do not re-tune the models. Final hyper-parameters are shown in Appendix~A.

\section{Experiments}

In Section~\ref{sec:sampling}, we compare several sampling methods for the \textsc{1-to-1} and \textsc{1-to-N} architectures. In Section~\ref{sec:ruleall} we then compare the MTL models with the single-task baselines. We turn to the issue of generalization in Section~\ref{sec:generalization}, where we use a recently introduced benchmark to evaluate the compositional generalization of our models. Finally, in Section~\ref{sec:rulesome} we report experiments between dataset pairs to find good auxiliary tasks.

\subsection{Task Sampling}
\label{sec:sampling}

\begin{table*}
\centering
\begin{tabular}{lcccccc}
\toprule
\textbf{Sampling} & \textbf{Geoquery} & \textbf{NLMaps} & \textbf{TOP} & \textbf{Overnight} & \textbf{AMR} & \textbf{Time}\\
\midrule
\textsc{Uniform} & 68.8 ($\mathsmaller{\pm} 3.8$) & 81.4 ($\mathsmaller{\pm} 2.6$) & 84.7 ($\mathsmaller{\pm} 0.1$) & 67.0 ($\mathsmaller{\pm} 0.8$) & 61.4 ($\mathsmaller{\pm} 1.7$) & 22h ($\mathsmaller{\pm} 2$h)\\
\textsc{Prop.} & 70.5 ($\mathsmaller{\pm} 1.9$) & 82.0 ($\mathsmaller{\pm} 0.5$) & 85.0 ($\mathsmaller{\pm} 0.0$) & 68.1 ($\mathsmaller{\pm} 0.4$) & 63.2 ($\mathsmaller{\pm} 0.4$) & 20h ($\mathsmaller{\pm} 2$h)\\
\textsc{LogProp.} & 70.7 ($\mathsmaller{\pm} 1.6$) & 82.8 ($\mathsmaller{\pm} 0.7$) & 85.2 ($\mathsmaller{\pm} 0.1$) & 68.3 ($\mathsmaller{\pm} 0.1$) & 62.9 ($\mathsmaller{\pm} 0.5$) & 18h($\mathsmaller{\pm} 4$h)\\
\textsc{SquareRoot} & 71.1 ($\mathsmaller{\pm} 2.5$) & 83.4 ($\mathsmaller{\pm} 1.1$) & 84.7 ($\mathsmaller{\pm} 0.0$) & 67.8 ($\mathsmaller{\pm} 0.6$) & 63.6 ($\mathsmaller{\pm} 1.0$) & 21h ($\mathsmaller{\pm} 4$h)\\
\textsc{Power} & 73.5 ($\mathsmaller{\pm} 1.4$) & 84.2 ($\mathsmaller{\pm} 0.5$) & 85.1 ($\mathsmaller{\pm} 0.3$) & 68.3 ($\mathsmaller{\pm} 0.4$) & 64.1 ($\mathsmaller{\pm} 0.3$) & 23h ($\mathsmaller{\pm} 7$h)\\
\textsc{Annealed} & 72.1 ($\mathsmaller{\pm} 0.0$) & 82.1 ($\mathsmaller{\pm} 0.2$) & 85.1 ($\mathsmaller{\pm} 0.3$) & 67.8 ($\mathsmaller{\pm} 0.2$) & 63.0 ($\mathsmaller{\pm} 0.6$) &  19h ($\mathsmaller{\pm} 2$h)\\
\textsc{Inverse} & 69.9 ($\mathsmaller{\pm} 2.4$) & 84.3 ($\mathsmaller{\pm} 0.8$) & 84.9 ($\mathsmaller{\pm} 0.2$) & 68.4 ($\mathsmaller{\pm} 0.7$) & 64.2 ($\mathsmaller{\pm} 0.7$) & 20h ($\mathsmaller{\pm} 2$h)\\
\textsc{Loss} & 73.3 ($\mathsmaller{\pm}1.9$) & 85.7 ($\mathsmaller{\pm}0.0$) & 85.2 ($\mathsmaller{\pm}0.1$) & 68.9 ($\mathsmaller{\pm}0.2$) & 64.2 ($\mathsmaller{\pm} 0.4$) & 15h ($\mathsmaller{\pm}2$h)\\
\bottomrule
\end{tabular}
\caption{Comparison of sampling strategies for the \textsc{1-to-N} architecture. We report the average over three runs with different random seeds. The standard deviation is in parentheses. All values reported are exact match, except for \textsc{AMR}, where \textsc{Smatch} is reported. We also report training times (in hours).}
\label{tab:sampling_n}
\end{table*}

\begin{table*}
\centering
\begin{tabular}{lcccccc}
\toprule
\textbf{Sampling} & \textbf{Geoquery} & \textbf{NLMaps} & \textbf{TOP} & \textbf{Overnight} & \textbf{AMR} & \textbf{Time}\\
\midrule
\textsc{Uniform} & 78.5 ($\mathsmaller{\pm} 1.4$) & 87.2 ($\mathsmaller{\pm} 0.2$) & 86.8 ($\mathsmaller{\pm} 0.2$) & 71.1 ($\mathsmaller{\pm} 0.2$) & 66.7 ($\mathsmaller{\pm} 0.5$) & 21h ($\mathsmaller{\pm} 4$h)\\
\textsc{Prop.} & 77.7 ($\mathsmaller{\pm}1.0$) & 86.2 ($\mathsmaller{\pm}0.2$) & 86.5 ($\mathsmaller{\pm}0.2$) & 70.6 ($\mathsmaller{\pm}0.2$) & 65.7 ($\mathsmaller{\pm}0.6$) & 16h ($\mathsmaller{\pm}1$h)\\
\textsc{LogProp.}  & 78.8 ($\mathsmaller{\pm}1.5$) & 87.2 ($\mathsmaller{\pm}0.1$) & 86.6 ($\mathsmaller{\pm}0.1$) & 71.0 ($\mathsmaller{\pm}0.3$) & 67.3 ($\mathsmaller{\pm}0.5$) & 23h ($\mathsmaller{\pm}3$h)\\
\textsc{SquareRoot} & 78.9 ($\mathsmaller{\pm} 1.5$) & 86.8 ($\mathsmaller{\pm} 0.1$) & 86.7 ($\mathsmaller{\pm} 0.2$) & 70.9 ($\mathsmaller{\pm} 0.0$) & 66.4 ($\mathsmaller{\pm} 0.3$) & 17h ($\mathsmaller{\pm} 0$h)\\
\textsc{Power} & 78.9 ($\mathsmaller{\pm} 0.6$) & 86.9 ($\mathsmaller{\pm} 0.3$) & 86.6 ($\mathsmaller{\pm} 0.1$) & 71.2 ($\mathsmaller{\pm} 0.6$) & 67.2 ($\mathsmaller{\pm} 0.5$) & 23h ($\mathsmaller{\pm} 2$h)\\
\textsc{Annealed} & 79.8 ($\mathsmaller{\pm} 0.7$) & 87.1 ($\mathsmaller{\pm} 0.1$) & 86.4 ($\mathsmaller{\pm} 0.2$) & 70.8 ($\mathsmaller{\pm} 0.4$) & 67.7 ($\mathsmaller{\pm} 0.3$) & 26h ($\mathsmaller{\pm} 1$h)\\
\textsc{Inverse} & 75.0 ($\mathsmaller{\pm} 2.3$) & 87.3 ($\mathsmaller{\pm} 0.4$) & 86.5 ($\mathsmaller{\pm} 0.1$) & 71.2 ($\mathsmaller{\pm} 0.5$) & 66.5 ($\mathsmaller{\pm} 0.7$) & 20h ($\mathsmaller{\pm} 3$h)\\
\textsc{Loss} & 76.5 ($\mathsmaller{\pm}1.4$) & 87.5 ($\mathsmaller{\pm} 0.2$) & 86.5 ($\mathsmaller{\pm} 0.1$) & 71.1 ($\mathsmaller{\pm} 0.1$) & 64.8 ($\mathsmaller{\pm}0.2$) & 11h ($\mathsmaller{\pm} 3$h)\\

\bottomrule
\end{tabular}
\caption{Comparison of sampling strategies for the \textsc{1-to-1} architecture.}
\label{tab:sampling_1}
\end{table*}



As discussed in Section~\ref{sec:seq2seq_mtl}, each training batch is sampled from one of the tasks. A simple sampling strategy is to pick the task uniformly, i.e., a training batch is extracted from task $t$ with probability $p_t = 1 / N$, where $N$ is the number of tasks. Due to the considerable differences in the sizes of our datasets, we further investigate the impact of previously proposed sampling strategies that take dataset sizes into account: 

\begin{itemize}
\item \textsc{Proportional} \cite{wang2019glue,sanh2019hierarchical}, where $p_t$ is proportional to the size of the training set of task $t$: $D_t$. That is: $p_t = D_t / (\sum_t{D_t})$;
\item \textsc{LogProportional} \cite{wang2019can}, where $p_t$ is proportional to $log(D_t)$; 
\item \textsc{SquareRoot} \cite{stickland2019bert}, where $p_t$ is proportional to $\sqrt{D_t}$;
\item \textsc{Power} \cite{wang2019can}, where $p_t$ is proportional to $D_t^{0.75}$;
\item \textsc{Annealed} \cite{stickland2019bert}, where $p_t$ is proportional to $D_t^{\alpha}$, with $\alpha$ decreasing at each epoch. When using proportional sampling methods, smaller tasks can be forgotten or interfered with, especially in the final epochs and when the final layers are shared \cite{stickland2019bert}. The method can therefore be particularly useful for the \textsc{1-to-1} architecture, where the decoder is shared.
\end{itemize}

We further test two additional sampling strategies:
\begin{itemize}
\item \textsc{Inverse}, where $p_t$ is proportional to $1/D_t$. The idea behind proportional sampling methods is to avoid overfitting smaller tasks and underfitting larger tasks. However, to the best of our knowledge, this intuitive hypothesis has not been explicitly tested. We test the opposite strategy.
\item \textsc{Loss}, where $p_t$ is proportional to $\mathcal{L}_t$, the loss on the development set for task $t$. This strategy therefore assigns higher sampling probabilities to harder tasks. This strategy is reminiscent of the active learning-inspired sampling method by \newcite{sharma2017learning}.
\end{itemize}



The results are shown in Table~\ref{tab:sampling_n} for \textsc{1-to-N} and in Table~\ref{tab:sampling_1} for \textsc{1-to-1}. 
We note that the choice of a sampling method depends on the MTL architecture and the dataset we want to optimize. The choice appears to be more critical for \textsc{1-to-N} than for \textsc{1-to-1}: for instance, in the case of \textsc{NLMaps}, the difference between the best sampling method and the worst is 4.3 for \textsc{1-to-N} and only 1.3 for \textsc{1-to-1}. This suggests that sampling methods are more relevant to train the dedicated layers. \textsc{1-to-1} appears to work well also with \textsc{Proportional}, which is expected to suffer for interference when sharing the final layers \cite{stickland2019bert}. As expected, \textsc{Annealed}, which explicitly addresses interference, works particularly well for \textsc{1-to-1}.

We presented \textsc{Inverse} as a way to test the intuition behind proportional strategies. Given the widespread use of proportional methods, we would expect \textsc{Proportional} to largely outperform \textsc{Uniform} and \textsc{Inverse}. We instead observe that in most cases it does not outperform \textsc{Inverse}, and in some cases underperforms it. For \textsc{1-to-1}, it does not even match the results of \textsc{Uniform}.
These results further suggest that there is not a generally superior sampling method, which should instead be picked as an additional hyper-parameter. They also highlight the need to further investigate sampling methods in MTL. The proposed \textsc{Loss} method is faster and performs particularly well for \textsc{1-to-N}. Henceforth, we use \textsc{Loss} for \textsc{1-to-N} and \textsc{Annealed} for \textsc{1-to-1}, which maximize the average accuracies across datasets. 

\subsection{One Semantic Parser to Parse Them All}
\label{sec:ruleall}

Table~\ref{tab:ruleall} compares the MTL results for the chosen sampling methods with the single-task baselines. We also report state-of-the-art parsing accuracies of each dataset for reference.
Note that \textsc{1-to-1} has more parameters than \textsc{1-to-N}. This is due to the fact that the increased sharing of \textsc{1-to-1} allowed us to train a larger model with 1024 hidden units instead of 512. 
In order to more directly compare the two MTL architectures, we also train a smaller \textsc{1-to-1} model (\textsc{1-to-1-Small}), which uses the same number of units as \textsc{1-to-N}. The results indicate that sharing also the decoder provides generally better results, even for the smaller model. Remarkably, compared to the single-task baseline, \textsc{1-to-1} achieves a 68\% reduction in the number of learnable parameters. Smaller models can have positive practical impacts as they decrease memory consumption hence reducing costs and carbon footprint \cite{schwartz2019allen}. We accomplish this without sacrificing parsing accuracies, which are competitive and in some cases higher than the baselines. This result is particularly promising, as we purposedly included a heterogeneous set of tasks and we use the same set of hyper-parameters for all of them. We can therefore train a single model with accurate parsing for a wide range of datasets, with fewer parameters.

\begin{table*}
\centering
\begin{tabular}{p{2.55cm}C{1.55cm}C{1.55cm}C{1.55cm}C{1.55cm}C{1.55cm}C{1.35cm}C{0.8cm}}
\toprule
\textbf{Model} & \textbf{Geoquery} & \textbf{NLMaps} & \textbf{TOP} & \textbf{Overnight} & \textbf{AMR} & \textbf{Time} & \textbf{Pars}\\
\midrule
\textsc{SOTA} & 89.0 & 64.4($\mathsmaller{\pm} 0.1$)$^{*}$ & 87.1 & 80.6$^{*}$ & 84.5 \\ 
\midrule
\textsc{Baseline} & 77.6($\mathsmaller{\pm} 2.2$) & 87.2($\mathsmaller{\pm} 0.7$) & 85.3($\mathsmaller{\pm} 0.4$) & 70.2($\mathsmaller{\pm} 0.9$) & 67.2($\mathsmaller{\pm} 0.3$) & 7h($\mathsmaller{\pm}0$h) & 721M\\
\textsc{1-to-N} & 73.3($\mathsmaller{\pm}1.9$) & 85.7($\mathsmaller{\pm}0.0$) & 85.2($\mathsmaller{\pm}0.1$) & 68.9($\mathsmaller{\pm}0.2$) & 64.2($\mathsmaller{\pm} 0.4$) & 15h($\mathsmaller{\pm}2$h) & 203M\\
\textsc{1-to-1} & 79.8($\mathsmaller{\pm} 0.7$) & 87.1($\mathsmaller{\pm} 0.1$) & 86.4($\mathsmaller{\pm} 0.2$) & 70.8($\mathsmaller{\pm} 0.4$) & 67.7($\mathsmaller{\pm} 0.3$) & 26h($\mathsmaller{\pm} 1$h) & 231M\\
\textsc{1-to-1-Small} & 76.7($\mathsmaller{\pm} 1.4$) & 85.0($\mathsmaller{\pm} 0.8$) & 85.9($\mathsmaller{\pm} 0.2$) & 69.7($\mathsmaller{\pm} 0.8$) & 64.9($\mathsmaller{\pm} 1.3$) & 20h($\mathsmaller{\pm} 5$h) & 169M\\
\bottomrule
\end{tabular}
\caption{Results of multitasking between all five datasets, compared to the baseline single-task parsers and state-of-the-art results (\textsc{SOTA}) on these datasets. \textsc{Pars} indicates the total number of parameters (in millions). Results marked with $^{*}$ are not directly comparable, as discussed in Section~\ref{sec:data}.}
\label{tab:ruleall}
\end{table*}

\subsection{Generalization}
\label{sec:generalization}

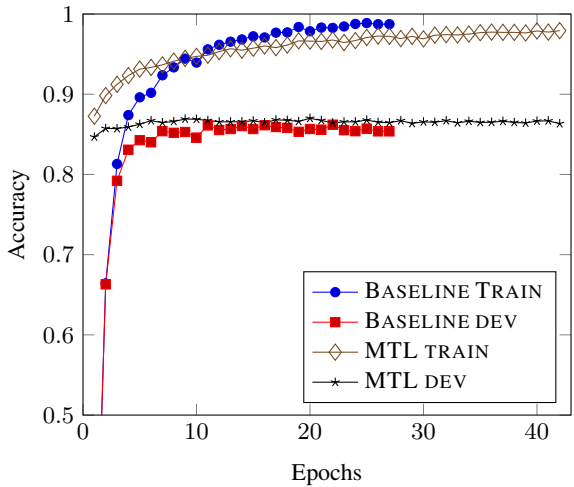
\begin{figure}[ht!]
{\small{
\begin{tikzpicture}
    \begin{axis}[
        xlabel=Epochs,
        width=8cm,
        ylabel=Accuracy,
        y label style={at={(axis description cs:0.06,0.5)}},
        xmin=0, xmax=43,
        ymin=0.5, ymax=1, 
        mark size=1.75pt,
        legend pos=south east,       
        legend cell align={left}
    ]

    \addplot plot coordinates {
(1, 0.19746568109820486)
(2, 0.6642027455121436)
(3, 0.8130939809926082)
(4, 0.8739880323829637)
(5, 0.8961633227736713)
(6, 0.9017951425554382)
(7, 0.9236184442097853)
(8, 0.9334741288278775)
(9, 0.944385779655051)
(10,    0.9394579373460049)
(11,    0.9560014079549455)
(12,    0.9616332277367124)
(13,    0.9655051038366772)
(14,    0.9683210137275607)
(15,    0.9721928898275255)
(16,    0.9707849348820837)
(17,    0.9767687434002112)
(18,    0.9771207321365716)
(19,    0.9838085181274199)
(20,    0.9785286870820133)
(21,    0.9831045406546991)
(22,    0.9827525519183387)
(23,    0.9845124956001408)
(24,    0.9876803942273847)
(25,    0.9887363604364661)
(26,    0.9873284054910243)
(27,    0.9873284054910243)
    }; 
    \addplot plot coordinates {
(1, 0.2048611111111111)
(2, 0.6629464285714286)
(3, 0.7921626984126984)
(4, 0.8308531746031746)
(5, 0.8427579365079365)
(6, 0.8400297619047619)
(7, 0.8539186507936508)
(8, 0.8519345238095238)
(9, 0.8529265873015873)
(10,    0.8459821428571429)
(11,    0.8611111111111112)
(12,    0.855406746031746)
(13,    0.8566468253968254)
(14,    0.8603670634920635)
(15,    0.8566468253968254)
(16,    0.861359126984127)
(17,    0.8591269841269841)
(18,    0.8581349206349206)
(19,    0.8531746031746031)
(20,    0.8566468253968254)
(21,    0.8559027777777778)
(22,    0.8621031746031746)
(23,    0.8551587301587301)
(24,    0.8539186507936508)
(25,    0.8571428571428571)
(26,    0.8536706349206349)
(27,    0.8536706349206349)
    };
    \addplot +[mark=diamond, mark options={scale=1.75}] plot coordinates {
(1, 0.872580077437522)
(2, 0.8982752551918338)
(3, 0.9123548046462513)
(4, 0.9232664554734249)
(5, 0.9313621964097148)
(6, 0.9334741288278775)
(7, 0.9366420274551215)
(8, 0.9412178810278071)
(9, 0.9447377683914114)
(10,    0.9472016895459345)
(11,    0.9486096444913763)
(12,    0.9531854980640619)
(13,    0.9567053854276664)
(14,    0.9549454417458642)
(15,    0.9574093629003871)
(16,    0.9602252727912707)
(17,    0.958113340373108)
(18,    0.961281239000352)
(19,    0.9665610700457585)
(20,    0.9662090813093981)
(21,    0.9658570925730376)
(22,    0.9676170362548399)
(23,    0.9644491376275959)
(24,    0.966913058782119)
(25,    0.9707849348820837)
(26,    0.9725448785638859)
(27,    0.9721928898275255)
(28,    0.9693769799366421)
(29,    0.9721928898275255)
(30,    0.9690249912002816)
(31,    0.9736008447729673)
(32,    0.9743048222456882)
(33,    0.9739528335093277)
(34,    0.975008799718409)
(35,    0.9767687434002112)
(36,    0.977472720872932)
(37,    0.9771207321365716)
(38,    0.9771207321365716)
(39,    0.9764167546638508)
(40,    0.9788806758183738)
(41,    0.977472720872932)
(42,    0.9792326645547342)
    };
    \addplot plot coordinates {
(1, 0.8467261904761905)
(2, 0.857390873015873)
(3, 0.8571428571428571)
(4, 0.8591269841269841)
(5, 0.8625992063492064)
(6, 0.8670634920634921)
(7, 0.8645833333333334)
(8, 0.8660714285714286)
(9, 0.8690476190476191)
(10,    0.8687996031746031)
(11,    0.8673115079365079)
(12,    0.8655753968253969)
(13,    0.8655753968253969)
(14,    0.8658234126984127)
(15,    0.8668154761904762)
(16,    0.8645833333333334)
(17,    0.8680555555555556)
(18,    0.8675595238095238)
(19,    0.8660714285714286)
(20,    0.8700396825396826)
(21,    0.8673115079365079)
(22,    0.8643353174603174)
(23,    0.8655753968253969)
(24,    0.8655753968253969)
(25,    0.8678075396825397)
(26,    0.8650793650793651)
(27,    0.8645833333333334)
(28,    0.8670634920634921)
(29,    0.8635912698412699)
(30,    0.8655753968253969)
(31,    0.8650793650793651)
(32,    0.8670634920634921)
(33,    0.8643353174603174)
(34,    0.8663194444444444)
(35,    0.8648313492063492)
(36,    0.8650793650793651)
(37,    0.8663194444444444)
(38,    0.8648313492063492)
(39,    0.8640873015873016)
(40,    0.8665674603174603)
(41,    0.8668154761904762)
(42,    0.863343253968254)
    };        
    \legend{\textsc{Baseline Train}\\\textsc{Baseline dev}\\\textsc{MTL train}\\\textsc{MTL dev}\\}
    \end{axis}
\end{tikzpicture}}}
\caption{Accuracies on training and dev split at each epoch for the \textsc{TOP} baseline and \textsc{1-to-1} MTL parser trained on all datasets and evaluated on \textsc{TOP}.}
\label{fig:plot_top}
\end{figure}

Table~\ref{tab:ruleall} also shows that MTL models are slower to converge. This is due to the regularization effect of training multiple tasks \cite{ruder2017overview}: as the loss on the development set keeps improving, the early stopping policy allows the MTL models to be trained for more epochs, resulting in longer training times. This regularization effect allows MTL to have better generalization \cite{caruana1997multitask,ruder2017overview}. In Figure~\ref{fig:plot_top} we compare the single-task \textsc{TOP} baseline against the \textsc{1-to-1} model trained on all datasets and evaluated on \textsc{TOP}. We show training and development accuracies as a function of the epochs. We observe that the baseline overfits earlier (early stopping is triggered earlier) and generalizes less (the gap between dev set and training set is larger) compared to the MTL model.

We further evaluate our models on the \textsc{CFQ} dataset \cite{keysers2019measuring}, designed to test compositional generalization. The idea behind datasets such as \textsc{CFQ} is to include test examples that contain unseen compositions of primitive elements (such as predicates, entities, and question types). To achieve this, a test set is sampled to maximize the compound divergence with the training set, hence containing unseen compositions (\textsc{MCD}). The dataset also contains a second test set, obtained with a random split. A parser that generalizes well is expected to achieve good results on both test sets. Table~\ref{tab:cfq} shows the results of our MTL model when adding \textsc{CFQ} as the sixth task.\footnote{For comparison with \newcite{keysers2019measuring}, we report mean and 95\%-confidence interval radius of 5 runs.}
We consider the relative improvements for \textsc{MCD} and \textsc{Random}, as the baseline values are considerably different. We note larger improvements on \textsc{MCD} (+27\%) than on \textsc{Random} (+13\%) when MTL is used. The results provide initial evidence that the MTL models result in better compositional generalization than the single-task baselines.
\begin{table}
\centering
\begin{tabular}{lcccccc}
\toprule
\textbf{Model} &\textbf{MCD} & \textbf{Random}\\
\midrule
\textsc{Keysers} & 17.9 ($\mathsmaller{\pm}0.9$) & 98.5 ($\mathsmaller{\pm}0.2$)\\
\midrule
\textsc{Baseline} & 
14.9 ($\pm 1.5$) & 84.9 ($\pm 0.7$)\\
\textsc{1-to-N} & 
16.8 ($\pm 0.6$) & 95.9 ($\pm 0.0$)\\
\textsc{1-to-1} & 
18.9 ($\pm 0.8$) & 95.6 ($\pm 0.1$)\\
\bottomrule
\end{tabular}
\caption{Results on the CFQ dataset. \textsc{Keysers} refers to the results reported by \newcite{keysers2019measuring} for the \textsc{Transformer} model. \textsc{MCD} reports the average of the three released \textsc{MCD} test sets.}
\label{tab:cfq}
\end{table}

\subsection{Auxiliary Tasks}
\label{sec:rulesome}
\begin{table*}
\centering
\begin{tabular}{llccccc}
\toprule
\multicolumn{2}{c}{\textbf{Model}} & \textbf{Geoquery} & \textbf{NLMaps} & \textbf{TOP} & \textbf{Overnight} & \textbf{AMR}\\
\midrule
\textsc{Baseline} & & 77.6($\mathsmaller{\pm} 2.2$) & 87.2($\mathsmaller{\pm} 0.7$) & 85.3($\mathsmaller{\pm} 0.4$) & 70.2($\mathsmaller{\pm} 0.9$) & 67.2($\mathsmaller{\pm} 0.3$)\\
\midrule
\multirow{ 5}{*}{\textsc{1-to-N}} 

& \textsc{+Geoquery} & N/A & 86.1 ($\mathsmaller{\pm}0.3$) & 85.8 ($\mathsmaller{\pm}0.0$) & 69.2 ($\mathsmaller{\pm}0.6$) & 63.3 ($\mathsmaller{\pm}2.1$)\\
& \textsc{+NLMaps} & 77.6 ($\mathsmaller{\pm}0.9$) & N/A & 85.6 ($\mathsmaller{\pm}0.1$) & 68.2 ($\mathsmaller{\pm}0.5$) & 64.4 ($\mathsmaller{\pm}0.5$)\\
& \textsc{+TOP} & 79.4 ($\mathsmaller{\pm}0.3$) & 83.0 ($\mathsmaller{\pm}1.0$) & N/A & 61.9 ($\mathsmaller{\pm}1.9$) & 65.3 ($\mathsmaller{\pm}0.5$)\\
& \textsc{+Overnight} & 75.7 ($\mathsmaller{\pm}0.8$) & 85.5 ($\mathsmaller{\pm}0.2$) & 85.0 ($\mathsmaller{\pm}0.3$) & N/A & 64.1 ($\mathsmaller{\pm}0.7$)\\
& \textsc{+AMR} & 82.0 ($\mathsmaller{\pm}0.4$) & 85.9 ($\mathsmaller{\pm}0.5$) & 85.8 ($\mathsmaller{\pm}0.1$) & 69.0 ($\mathsmaller{\pm}0.4$) & N/A\\

\midrule
\multirow{ 5}{*}{\textsc{1-to-1}}
& \textsc{+Geoquery} & N/A & 87.4 ($\mathsmaller{\pm}0.6$) & 86.5 ($\mathsmaller{\pm}0.1$) & 70.9 ($\mathsmaller{\pm}0.8$) & 66.3 ($\mathsmaller{\pm}0.3$)\\
& \textsc{+NLMaps} & 80.0 ($\mathsmaller{\pm}1.8$) & N/A & 86.4 ($\mathsmaller{\pm}0.3$) & 69.7 ($\mathsmaller{\pm}1.6$) & 67.3 ($\mathsmaller{\pm}0.1$)\\ 
& \textsc{+TOP} & 80.5 ($\mathsmaller{\pm}1.5$) & 85.4 ($\mathsmaller{\pm}0.5$) & N/A & 65.8 ($\mathsmaller{\pm}1.2$) & 66.8 ($\mathsmaller{\pm}0.5$)\\
& \textsc{+Overnight} & 77.3 ($\mathsmaller{\pm}1.5$) & 87.0 ($\mathsmaller{\pm}0.3$) & 86.2 ($\mathsmaller{\pm}0.4$) & N/A & 67.0 ($\mathsmaller{\pm}0.3$)\\
& \textsc{+AMR} & 77.7 ($\mathsmaller{\pm}0.2$) & 86.9 ($\mathsmaller{\pm}0.3$) & 86.7 ($\mathsmaller{\pm}0.3$) & 70.9 ($\mathsmaller{\pm}0.1$) & N/A\\
\bottomrule
\end{tabular}
\caption{Experiments on dataset pairs. The rows are the auxiliary tasks and the columns are the main tasks.}
\label{tab:rulesome}
\end{table*}

Finally, we trained MTL models on dataset pairs to find what datasets are good auxiliary tasks (i.e., tasks that are helpful to other tasks). Note that we do not tune the hyper-parameters of each pairwise model, as we would need to do a costly hyper-parameter search over 40 models. The results are shown in Table~\ref{tab:rulesome}. The problem of choosing auxiliary tasks has been shown to be challenging \citep{alonso2016multitask,bingel2017identifying,hershcovich2018multitask}. Similar to task sampling methods, there is not an easy recipe to choose the auxiliary tasks. However, our results elicit the following surprising observations: 
\begin{enumerate}
  \item \textsc{AMR} is the only dataset to use graph-structured MRL, due to explicit representation of pronominal coreference, coordination, and control structures. It is also the only non-Q\&A dataset. Nevertheless, we note that \textsc{AMR} is a competitive auxiliary task, possibly due to its large size and scope.
  It is also surprising that \textsc{AMR} is often more helpful in the \textsc{1-to-1} setup, where the whole network is shared and more related tasks are expected to be preferred.


  \item Transfer learning is often used to provide low-resource tasks with additional data from a higher-resource task. However, in our experiments, \textsc{Geoquery}, our smallest dataset, appears to be helpful for the larger \textsc{TOP} dataset.


\end{enumerate}

\section{Related Work}
\label{sec:relatedwork}

A number of alternative meaning representations and semantic parsing datasets have been developed in recent years, spanning from broad-range meaning representations such as Parallel Meaning Bank \cite{abzianidze2017parallel} and UCCA \cite{abend2013universal}, to domain-specific datasets such as LCQUAD \cite{dubey2019lc} and KQA Pro \cite{shi2020kqa}.

Following previous work on semantic parsing \cite{jia2016data,konstas2017neural,fan2017transfer,hershcovich2018multitask,rongali2020don}, the baseline parser used in this work is based on the popular attentive sequence to sequence framework \cite{sutskever2014sequence,bahdanau2015neural}. Pointer networks \cite{vinyals2015pointer} have demonstrated the importance of decoupling the job of generating new output tokens from that of copying tokens from the input. To achieve this, our models use copy mechanisms, following previous work on semantic parsing \cite{rongali2020don}. We further rely on pre-trained embeddings \cite{liu2019roberta}.

Compositional generalization has recently attracted attention \cite{neyshabur2017exploring,lake2018generalization,finegan2018improving,hupkes2018learning,keysers2019measuring}. We used the CFQ dataset \cite{keysers2019measuring}, with the purpose of assessing their compositional generalization. 
MTL \cite{caruana1997multitask,ruder2017overview} based on sequence to sequence models has been used to address several NLP problems such as syntactic parsing \cite{luong2016multi} and Machine Translation \cite{dong2015multi,luong2016multi}.
For the task of semantic parsing, MTL has been employed as a way to transfer learning between domains \citep{damonte2019practical} and datasets \citep{fan2017transfer,lindemann2019compositional,hershcovich2018multitask,lindemann2019compositional}. A shared task on multi-framework semantic parsing with a particular focus on MTL has been recently introduced \citep{oepen2019mrp}.
The \textsc{1-to-N} and \textsc{1-to-1} models have been previously experimented with by \newcite{fan2017transfer}, with the latter being an MTL variant of the models used for multilingual parsing by \newcite{johnson2017google}. An alternative to MTL for transfer learning is based on pre-training on a task and fine-tuning on related tasks \cite{thrun1996learning}. It has been investigated mostly for machine translation tasks \cite{zoph2016transfer,johnson2017google,bansal2019pre} but also for semantic parsing \cite{damonte2019practical}.

\section{Conclusions}

We used MTL to train joint models for a wide range of semantic parsing datasets.
We showed that MTL provides large parameter count reduction while maintaining competitive parsing accuracies, even for inherently different datasets. We further discussed how generalization is another advantage of MTL and we used the CFQ dataset to suggest that MTL achieves better compositional generalization. We leave it to future work to further investigate this type of generalization in the context of MTL. 
We compared several sampling methods, indicating that proportional sampling is not always optimal, showing room for improvements, and introducing a loss-based sampling method as a competitive and promising alternative. 
We were surprised to see the positive impact of low-resource (\textsc{Geoquery}) and less-related (\textsc{AMR}) datasets can have as auxiliary tasks.
Challenges in finding optimal sampling strategies and auxiliary tasks suggest that they should be treated as hyper-parameters to be tuned.

\section*{Acknowledgments}

The authors would like to thank the three anonymous reviewers for their comments and the Amazon Alexa AI team members for their feedback.

\bibliographystyle{acl_natbib}
\bibliography{anthology,acl2021}

\appendix

\section{Hyper-parameters}
\label{sec:appendix_hp}
Table~\ref{tab:appendix_hp} reports the final hyper-parameters used for our experiments.

\begin{table*}
\centering
\begin{tabular}{lccccccccc}
\toprule
\multicolumn{2}{c}{\textbf{Model}} & \textbf{Batch} & \textbf{lr} & \textbf{Layers} & \textbf{Units} &  \textbf{Heads} & \textbf{Dropout} \\
\midrule
\multirow{ 5}{*}{\textsc{Baseline}} 
& \textsc{Geoquery} & 100 & 0.05 & 3 & 512 & 4 & 0.1 \\
& \textsc{NLMaps} & 50 & 0.05 & 4 & 512 & 16 & 0.05\\
& \textsc{TOP} & 200 & 0.05 & 3 & 512 & 4 & 0.04 \\
& \textsc{Overnight} & 10 & 0.05 & 3 & 700 & 4 & 0.03 \\
& \textsc{AMR} & 10 & 0.05 & 4 & 512 & 4 & 0.03 \\
\midrule
\textsc{1-to-N} & & 10 & 0.05 & 3 & 512 & 4 & 0.1 \\
\textsc{1-to-1} & & 10 & 0.05 & 3 & 1024 & 4 & 0.1 \\
\textsc{1-to-1-Small} & & 10 & 0.05 & 3 & 512 & 4 & 0.1 \\
\bottomrule
\end{tabular}
\caption{Hyper-parameter selected for baselines and MTL models. From left to right the hyper-parameters are: batch size, learning rate, number of layers and units in the decoder, number of attention heads, and dropout ratio.}
\label{tab:appendix_hp}
\end{table*}

\end{document}